\documentclass[runningheads]{llncs}

\usepackage[T1]{fontenc}
\usepackage{graphicx}
\usepackage{amsmath}
\usepackage{amsfonts}
\usepackage{booktabs}
\usepackage{multirow}
\usepackage{algorithm}
\usepackage{algorithmic}
\usepackage{cleveref}

\makeatletter
\renewcommand{\subsection}{\@startsection{subsection}{2}{\z@}%
  {-1.2ex plus -0.3ex minus -0.2ex}
  {0.6ex plus 0.2ex}
  {\normalfont\normalsize\bfseries}}
\makeatother
\usepackage[font=small,skip=4pt]{caption}

\newcommand{\concat}{\ ^\frown}

\begin{document}
\title{COAgents: Multi-Agent Framework to Learn and Navigate Routing Problems Search Space}

\author{Oleksandr Yakovenko\inst{1} \and
        Mahdi Mostajabdaveh\inst{1} \and
        Cheikh Ahmed\inst{1} \and
        Abdullah Ali Sivas\inst{1} \and
        Xiaorui Li\inst{1} \and
        Zirui Zhou\inst{1} \and
        Mao Kun\inst{2}}

\authorrunning{O. Yakovenko et al.}
%
\institute{Huawei Technologies Canada, 4321 Still Creek Dr, Burnaby, BC, Canada V5C 6S7 \and
Huawei Technologies, China}
\maketitle              
%
\begin{abstract}
Although Vehicle Routing Problems (VRP) are essential to many real-world systems, they remain computationally intractable at scale due to their combinatorial complexity. Traditional heuristics rely on handcrafted rules for local improvements and occasional \textit{jumps} to escape local minima, but often struggle to generalize across diverse instances. We introduce \textbf{COAgents}, a cooperative multi-agent framework that models the search process as a graph: nodes represent solutions, and edges correspond to either local refinements or large perturbations for diversification (i.e., jumps). A \textit{Partial Search Graph} (PSG) is dynamically constructed during search, enabling COAgents to train a Node Selection Agent and a Move Selection Agent to guide intensification, and a Jump Agent to trigger well-timed explorations of new regions. Unlike end-to-end learning approaches, COAgents cleanly separates problem-agnostic search control from compact domain-specific encoding, facilitating adaptability across tasks. 
Extensive experiments on the CVRP and VRPTW benchmarks show that COAgents remains competitive with several learn-to-search baselines on CVRP and sets a new state of the art among learning-based methods on the more challenging VRPTW instances, reducing the gap to the best-known solutions by 14\% at $N\!=\!100$ and 44\% at $N\!=\!50$ relative to the strongest neural solver (POMO), and by 21\% and 40\% respectively relative to ALNS.

Code is available at \url{https://github.com/mahdims/COAgents}.

\end{abstract}

\section{Introduction}
Combinatorial optimization problems are central to many real-world decision-making scenarios, including finance, e-commerce, logistics, and manufacturing~\cite{Petropoulos03032024}. Despite breakthroughs in generic solvers, the inherently discrete nature of these problems means that even state-of-the-art solvers often fall short on practical-sized instances, particularly in real-world large-scale logistics and routing applications~\cite{mostajabdaveh2025branch}. Consequently, there is a demand for fast, reliable heuristic methods that can produce high-quality solutions within limited computation times. Numerous variants of the Vehicle Routing Problem (VRP) offer a solid yet diverse playground for prototyping and testing such OR methods.

Many traditional heuristics for VRPs are built on iterative local search: at each step, they (i) select a current solution, (ii) apply one or more neighborhood moves, and (iii) accept or reject the resulting solution via a heuristic rule. To avoid getting stuck, the VRP solvers augment this process with diversification, randomized constructions, or large perturbations that “jump” to new regions. The local search algorithms are repeatedly answering three core questions:  
1) \textbf{Which solution} becomes the next incumbent?  
2) \textbf{Which operator} should be applied next?  
3) \textbf{When and how} should the search be redirected to escape a local minimum?  

Despite their widespread success, classic methods rely on handcrafted, problem-specific rules for each decision, demanding extensive domain expertise and costly trial-and-error tuning \cite{khalil2017learning}. Even minor shifts in problem formulation or input data can invalidate the tuned parameters, forcing expert intervention and comprehensive re-engineering. Moreover, static rule sets cannot accumulate or adapt from past search experience, nor adjust online to evolving instance distributions. This brittleness and maintenance burden motivate our shift to the \textbf{COAgents} framework, where learned agents replace fixed heuristics to deliver adaptive, data‐driven decision policies with minimal manual tuning. 

\begin{figure*}[htp!]
\centering
\includegraphics[width=\textwidth]{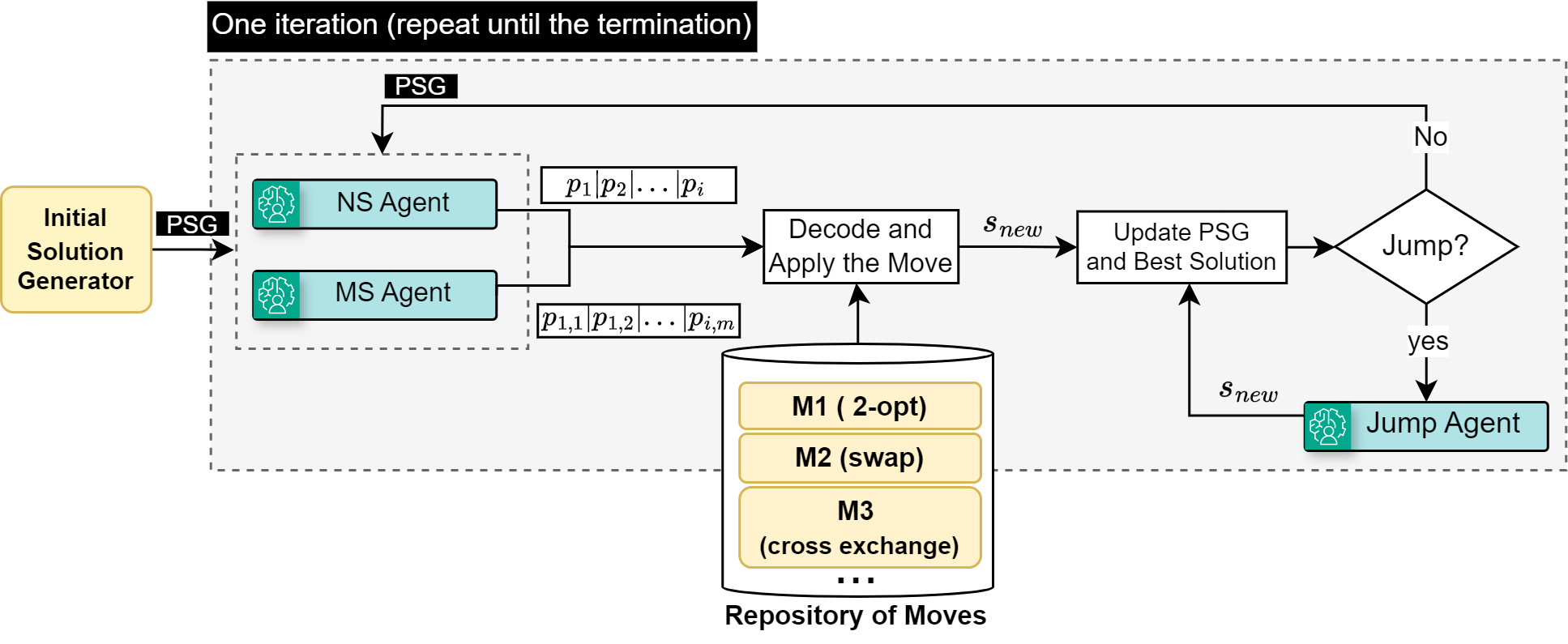}
\caption{The COAgents which defines how our three agents are collaborating to orchestrate the search.}
\label{fig:algMain}
\end{figure*}

{COAgents} is a general multi‐agent framework that leverages search history to orchestrate local improvement heuristics via three learned agents, the \emph{Node Selection Agent} (NSA), the \emph{Move Selection Agent} (MSA), and the \emph{Jump Agent} (JA), operating over a \emph{Partial Search Graph} (PSG). A PSG is the visited subgraph of the entire search space, with nodes representing explored solutions and edges representing moves or jumps, thereby encoding the search history. Starting from an initial solution, NSA selects a candidate node and MSA predicts the most promising improvement heuristic, driving local exploration. When repeated failed attempts signal stagnation, JA generates a new solution from scratch based on the search history, steering the search into previously unexplored or unreachable regions. This jump transformation is not a standard operator but a learned restart that escapes local minima and resumes downhill search from a fresh point. Control then returns to NSA and MSA for renewed local search. Figure~\ref{fig:algMain} outlines this top‑level loop: NSA, MSA, move application, PSG update, and conditional invocation of JA. The loop is executed until the computational budget (for example, time limits or number of explored nodes) is reached. 
 
\paragraph{Contributions}  
COAgents is the first multi-agent framework that models and learns a combinatorial search space of VRPs via collaborative decision-making over a Partial Search Graph. Our key contributions are:  
(1) \textbf{Novel learned agents.} We introduce two new agents, the Node Selection Agent, which prioritizes promising regions of the search, and the Jump Agent, which predicts learned restarts to escape stagnation—alongside a learned Move Selection Agent, collectively replacing handcrafted rules.  
(2) \textbf{Unified, weight-shared architecture.} Agents share a common neural backbone, yielding data-efficient training, easy transfer to new problem variants, and reduced expert tuning.  
(3) \textbf{Partial Search Graph representation.} We formalize the PSG as a general, history-aware state encoding that drives agent decisions. 
(4) \textbf{Solver‐ and problem‐adaptive framework.} COAgents wraps around a existing local search routines and neural solvers to enhance them with adaptive decision policies that seamlessly adapts to different variants of VRP problems.
Empirically, COAgents establishes a new state of the art among learning-based methods on VRPTW, reducing the gap to the best-known solutions by up to 44\% over the strongest neural baseline.

\section{Related Works}

\textbf{Hyper-heuristics} are automated methods for selecting or generating heuristics to solve combinatorial problems, first introduced in the 1990s~\cite{denzinger1996high}. Interest in hyper-heuristics has grown due to their potential to reduce development overhead and advances in machine learning~\cite{drake2020recent}. They are broadly classified into \emph{selection} and \emph{generation} approaches~\cite{burke2019classification}. Selection hyper-heuristics choose from a set of low-level heuristics to apply to the current solution and then decide whether to accept the new solution~\cite{hyper_heuristics_EJOR}. Generation hyper-heuristics, by contrast, create new heuristics by combining building blocks or identifying patterns in existing ones \cite{reevo}. COAgents differs from these methods as it is a multi‐agent framework that not only selects heuristics but also makes decisions on solution selection and diversification jumps.   

\textbf{Learn-to-Search methods.}  
\cite{lu2019learning} introduced a learn-to-search (L2S) framework for VRPs that iteratively improves an initial solution by selecting customized operators through a reinforcement learning (RL) controller. Building on this idea, \cite{chen2019learning} proposed a deep RL approach that iteratively modifies local solution components until convergence, successfully addressing problems such as the Capacitated Vehicle Routing Problem (CVRP), scheduling, and  Traveling Salesperson Problem (TSP). More recently, \cite{lagos2024multi} formulated operator selection in hyper-heuristics as a Multi-Armed Bandit (MAB) problem, leveraging Thompson Sampling and EXP3 to adaptively handle adversarial and non-stationary settings. These online MAB algorithms dynamically tune parameters during runtime, and were tested on the Vehicle Routing Problem with Time Windows (VRPTW). COAgents differs fundamentally from these methods by going beyond operator selection. It introduces a multi-agent framework that simultaneously decides which solution to expand, which operator to apply, and when and where to diversify, while explicitly modeling the search space as a graph for richer context.

\textbf{ALNS operator selection.}  
Recent studies have applied Deep Reinforcement Learning (DRL) to improve operator selection in Adaptive Large Neighborhood Search (ALNS). \cite{ALNS2020Neural} and \cite{kallestad2023general} used multi-layer perceptron architectures, relying only on high-level search features (e.g., iteration count, temperature, and optimality gap) and ignoring solution-specific details. This omission may limit operator-selection effectiveness. To address this, \cite{johnn2024graph} use a Graph Neural Network (GNN) that embeds the current solution graph to guide operator selection from a large pool (28 destroy and 7 repair operators), achieving consistent gains across five routing problems. Building on these ideas, \cite{reijnen2024online} introduced DR-ALNS, a DRL framework that jointly learns operator selection and dynamically tunes algorithm parameters. Evaluated on the orienteering problem, DR-ALNS outperformed both traditional and Bayesian-tuned ALNS of \cite{SMAC3}. Unlike these approaches, COAgents explicitly models the entire search trajectory as a PSG, enabling richer context for its solution selection, operator selection, and jump agents.

\section{Problem Statement}
\label{sec:problem_statement}
We define the \textit{combinatorial solution space} of a VRP variant as a directed graph $G^{CO}_{P}(\mathcal{S}, E)$ where each node $s \in \mathcal{S}$ represents a distinct solution to the VRP problem $P$ and $E$ is a collection of \textit{neighbourhood moves} for the problem $P$. An edge $e_{ij} \in E$ connects two nodes $s^i$ and $s^j$ if there exists a move that converts $s^i$ into $s^j$. An example of a move that corresponds to an edge is the 2-opt heuristic for the Traveling Salesman (Sub-)Problem.

Any global optimum $s^\ast$ of the problem P is a node in the graph $G^{CO}_{P}(\mathcal{S}, E)$. However, there is no guarantee that an arbitrary solution $s$ can be transformed into $s^\ast$ via a sequence of moves. Equivalently, $G^{CO}_{P}(\mathcal{S}, E)$ may be a disconnected graph. Additionally, for each local optimal solution $s^{\ast}_i$, there exists an induced subgraph $G^{CO}_{P}(s^{\ast}_i;\mathcal{S}, E)$ such that if $s$ is a node of $G^{CO}_{P}(s^{\ast}_i;\mathcal{S}, E)$ then sequences of moves from $s$ are likely to lead to $s^{\ast}_i$. We call these induced subgraphs \textit{basins of attraction}. Hence, the graph traversal is biased, and neighborhood moves may lead to a local optimum even when a path to a global optimum exists.

In practical scenarios, only a subset of all possible moves are available and/or cheap. Therefore, we introduce the \textit{search space} as the subgraph $G^{SS}_P(\mathcal{S}, E_{\mathcal{M}}) \subset G^{CO}_P(\mathcal{S}, E)$, where $\mathcal{M}$ denotes the set of available moves, and $E_{\mathcal{M}}$ is the corresponding set of edges. From now on, we refer to these spaces as $G^{SS}$ and $G^{CO}$, respectively. Observe that edges of $G^{SS}$ are a subset of edges of $G^{CO}$ and the optimal solution $s^\ast$ is a node in $G^{SS}$. 

\begin{figure}[htp]
  \centering

  \begin{minipage}{0.48\columnwidth}
    \centering
    \includegraphics[width=\linewidth]{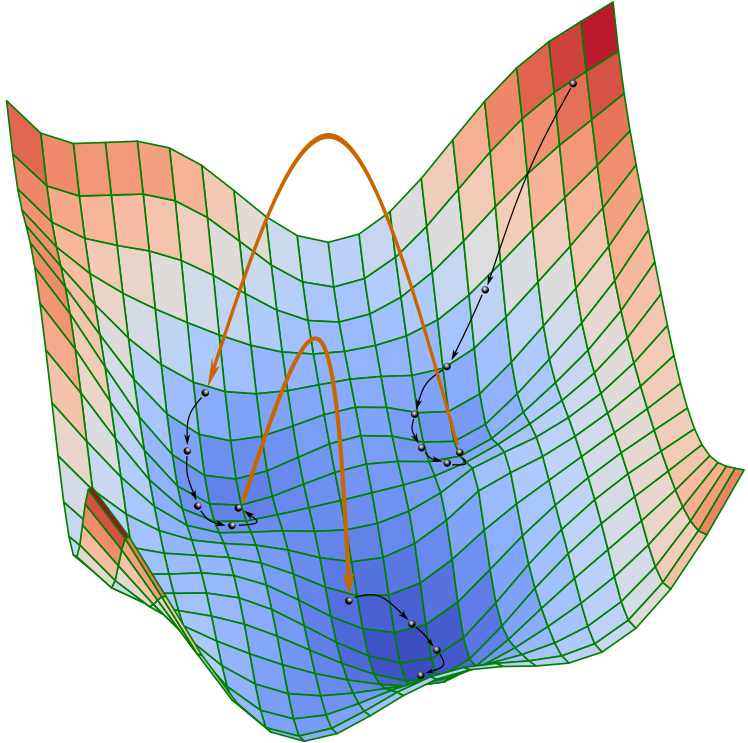}
    \caption{\textit{Moves} (black) and \textit{jumps} (orange) work together in the search of the global optimum.}
    \label{fig:PES}
  \end{minipage}
  \hfill
  \begin{minipage}{0.48\columnwidth}
    \centering
    \includegraphics[width=\linewidth]{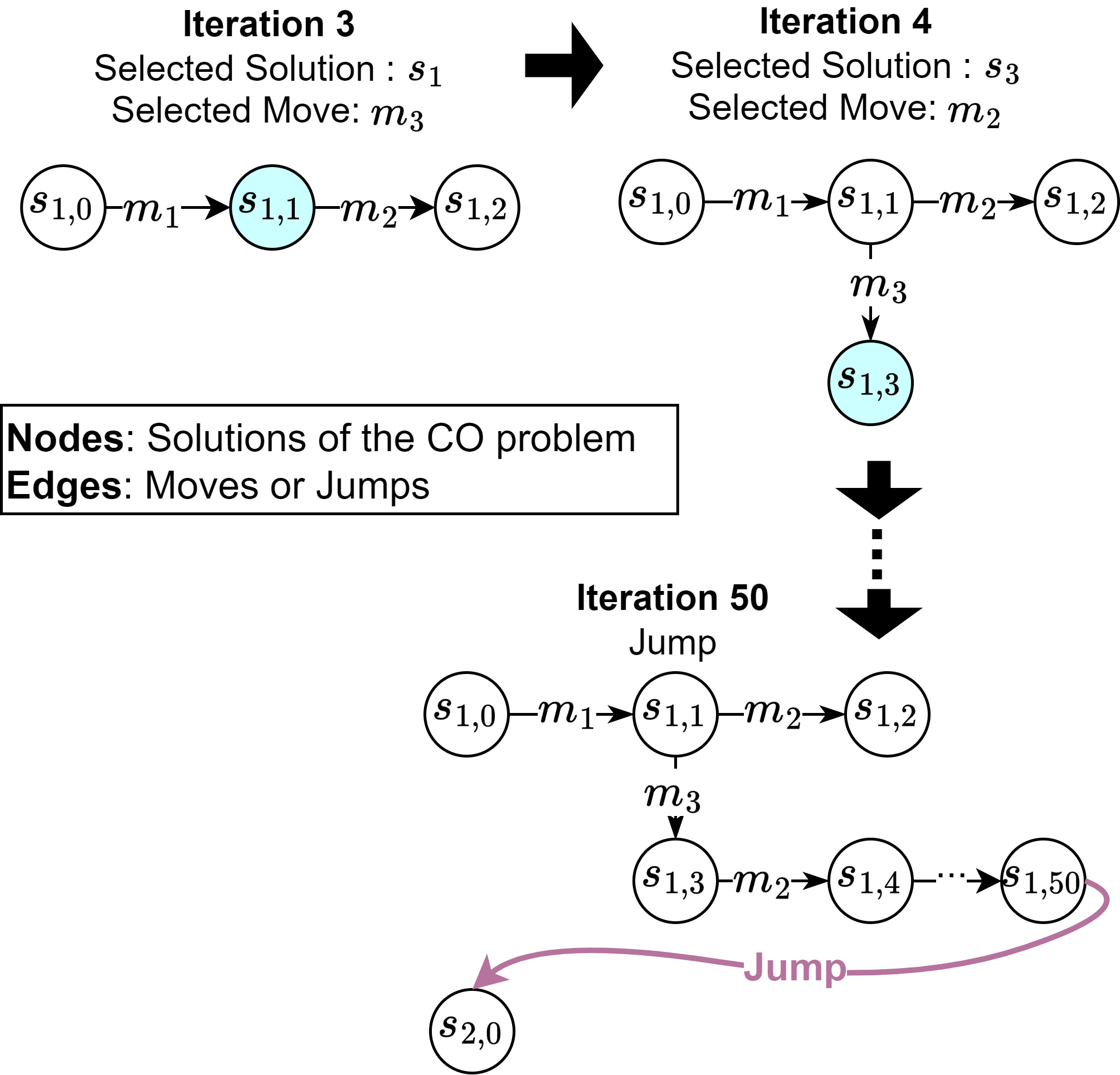}
    \caption{Evolution of a PSG during the algorithm. \(s\) denotes solutions and \(m\) are moves.}
    \label{fig:PSG}
  \end{minipage}

\end{figure}

Now, we cast the problem of finding an optimal solution $s^{\ast}\in\mathcal{S}$ of a given VRP problem $P$ to 
\begin{itemize}
    \item finding a set of edges $E_J$, not necessarily corresponding to any legitimate move, such that $G^{SS}_P(\mathcal{M}; \mathcal{S}, E_{\mathcal{M}} \bigcup E_{J}))$ is connected,
    \item and finding the shortest path between an initial solution $s^{(0)}\in\mathcal{S}$ and $s^{\ast}\in\mathcal{S}$ over the graph $G^{SS}_P(\mathcal{M}; \mathcal{S}, E_{\mathcal{M}} \bigcup E_{J}))$.
\end{itemize}
We name the set $E_{J}$ \textit{jumps}. Note that jumps do not have to improve the objective value; hence, they are not affected by basins of attraction. See \cref{fig:PES} for an illustration. 

Explicit representation of $G^{SS}$ is not available due to space and time constraints. Instead, we will aim to train our model to implicitly construct the search space during inference. We achieve this by constructing a set of \textit{Partial Search Graphs} and using this set as an additional input to the model. A partial search graph is an \textit{explored} subspace of the entire search space. Nodes represent solutions encountered during the search process. Edges represent either the moves or the jumps. As a result, a PSG captures a trajectory of the search within the search space~$G^{SS}$, showing the parts of the space that have been visited and how they are connected. \cref{fig:PSG} demonstrates how a PSG evolves between iterations. 

In this study, we propose a method that learns how to guide the exploration of the search space from previous experiences and how to generate \textit{jumps} during exploration. We will use PSGs to identify patterns in solution exploration and to guide further exploration, improving search efficiency by focusing on unexplored or promising regions. We discuss our approach in the \textit{Methods} section.

\section{Solution Method: COAgents} \label{sec:method}

COAgents consists of three agents—the NSA, MSA, and JA, which interact as shown in Fig.~\ref{fig:algMain}. Each agent is implemented as a graph-neural-network–based policy trained to maximize long-term progress toward the optimum. In this section, we discuss the data representation, model architecture, and training procedures for these three agents.

\subsection{Search History Data Representation} \label{ss:search_embedding}

Navigating the non-convex combinatorial solution space of VRPs is a challenging task. To enhance agents' understanding of the solution space, it is crucial to integrate information about the problem, the solution, and the exploration history. While problem and solution encodings are inherently problem-specific, the history exhibits a consistent structure across many variants of combinatorial optimization problems. Next, we discuss the problem-independent representation of the exploration history in detail.

\paragraph{Partial Search Graph} 
The presented framework is built upon the PSG data structure, which represents the search history. As described above, the PSG is essentially a directed, disjointed graph. 
Each node in the PSG corresponds to a single solution $s_i$ of a combinatorial optimization problem, and each directed edge $(s_i \to s_j)$ indicates that $s_j$ was reached by applying a neighborhood move operator to $s_i$.
The PSG naturally decomposes into connected components—which we call \emph{samples}—each of which explores the neighborhood of a local optimum $s^\ast_k$. In turn, each sample is a flattened collection of embeddings of individual solutions $s_i$ together with their connectivity information. 

\paragraph{Node Features} 
 Each solution embedding is represented by a fixed-size vector \(\mathbf{u_i}\) which encodes global solution-specific attributes, including objective value, feasibility, and graph-level aggregates such as the gap relative to its parent in PSG, the number of derived solutions and their objectives, etc. Additionally, each solution is accompanied by a dynamically sized matrix \(\mathbf{V_i}\) (which varies across batches) that encapsulates its internal structure. Both \(u_i\) and \(V_i\) are constructed by  concatenating the \textit{structural} and \textit{positional} embeddings of a CO solution \(s_i\). The detailed computation of these embeddings is discussed next, while the complete set of per-problem features utilized in our experiments is provided in
 \textbf{Appendix A}

\paragraph{Input Embeddings} Our agents operate on node embeddings of the PSG introduced above.  Since all PSG edges (moves and jumps) are provided explicitly, we compute each node’s positional embedding \(u_{\mathrm{pos}}\) via a simple random-walk encoder \cite{PEs}.  The structural embedding of the local and global features is then obtained by

\begin{equation}
\begin{split}
u_{\mathrm{str}} &= \mathrm{Linear}(x_{u}) \;\in \mathbb{R}^{d_u}, \\
V_{\mathrm{str}} &= \mathrm{Linear}(x_{V}) \;\in \mathbb{R}^{d_V}.
\label{eq:structural}
\end{split}
\end{equation}

Each solution also carries a problem-specific positional embedding \(V_{\mathrm{pos}}\); details for its construction in each domain are given in
\textbf{Appendix B}. Finally, we concatenate structural and positional embeddings:
\begin{equation}
\begin{aligned}
u_{inp} = u_{\mathrm{str}} \concat u_{\mathrm{pos}},\quad V_{inp} = V_{\mathrm{str}} \concat V_{\mathrm{pos}}.
\end{aligned}
\label{eq:concat}
\end{equation}
where \(u_{inp}\in\mathbb{R}^{\,d_u + d_{\mathrm{pos}}}\) and \(V_{inp}\in\mathbb{R}^{\,d_V + d'_{\mathrm{pos}}}\).  

\subsection{Agent Architectures}
\label{ss:agents}

\begin{figure*}[htp]
    \centering
    \includegraphics[width=\textwidth]{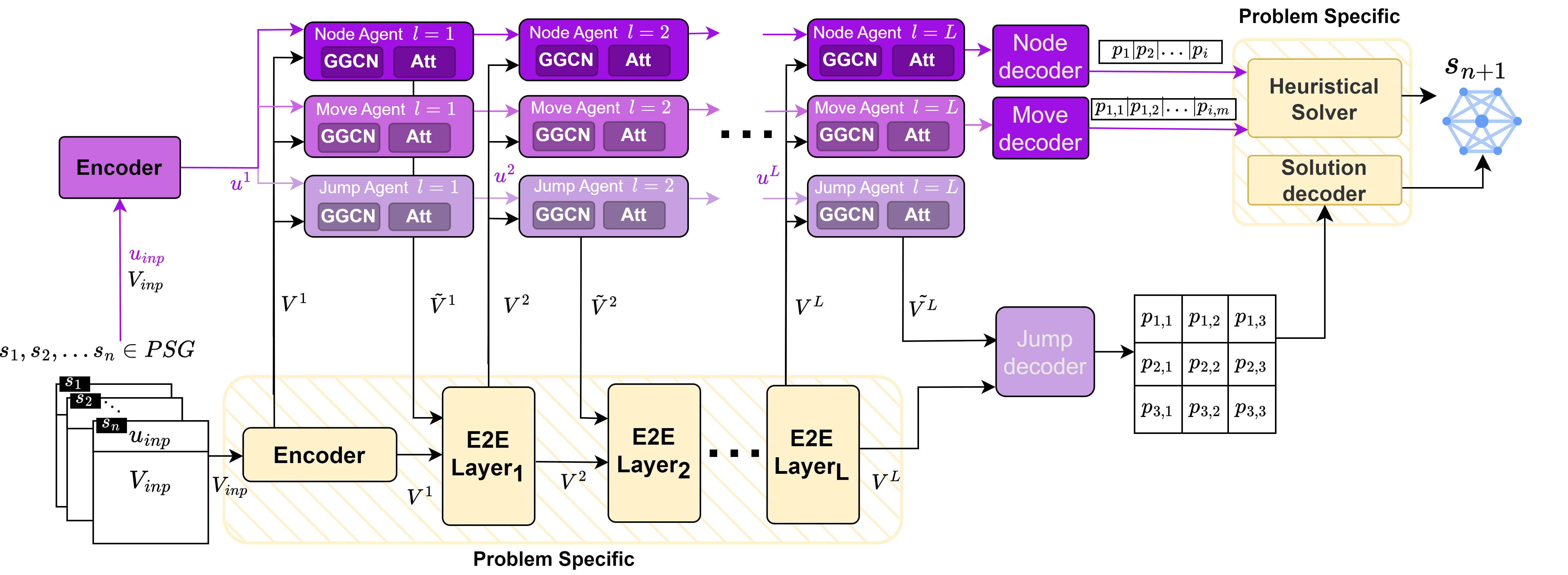}
    \caption{Agents model architecture}
    \label{fig:architecture}
\end{figure*}

All the agents are composed of the same core blocks, with only minor differences in functionality. We employ a combination of Gated Graph Convolution (GGCN), Transformer, and problem-specific modules as the universal core block of the agents' networks. GGCN layers propagate and filter messages among each node’s immediate neighbors in PSG to capture local structural patterns, while Transformer layers apply global self-attention across all nodes to model long-range dependencies. Given GGCN's graph-specialized capabilities, we opted for a standard Transformer architecture rather than a graph attention mechanism. Figure~\ref{fig:architecture} illustrates the general architecture of agents. 

Our network comprises a stack of interleaved \textit{CoreBlocks}, each consisting of a Gated Graph Convolutional Network (GGCN), followed by a Transformer layer, and finally an augmented problem-specific layer. Let \(u_i^{(l-1)}\in\mathbb{R}^{d_u}\) be the global embedding of node \(i\) and let \(v_{ik}^{(l-1)}\in\mathbb{R}^{d_v}\) denote the \(k\)-th row of its internal structure matrix \(V_{ik}^{(l-1)} = [v_{0,0}^{l-1}, v_{0,1}^{l-1}, \ldots, v_{0,k}^{l-1}, v_{1,0}^{l-1}, \ldots, v_{M,N}^{l-1}]\) where \(N\) is the size of the problem and \(M\) is the amount of trial solutions in the scope of an agent. The CoreBlock transformation is then: 

\begin{equation}
\begin{split}
(\hat{u}_i^{(l)},\,\hat{V}_i^{(l)}) &= \mathrm{GGCN}(u_i^{(l-1)},\,V_i^{(l-1)},\,E), \\
(\tilde{u}_i^{(l)},\tilde{V}_i^{(l)}) &= \mathrm{Transformer}(\hat{u}_i^{(l)},\hat{V}_i^{(l)}), \\
\bar{V}_{i}^{(l)} &= \mathrm{E2E}(V_i^{(l-1)},\tilde{V}_i^{(l)}), \\
u_i^{(l)} &= u_i^{(l-1)} + \tilde{u}_i^{(l)}, \quad
V_i^{(l)} = V_i^{(l-1)} + \bar{V}_{i}^{(l)}.
\end{split}
\label{eq:e2e_integration}
\end{equation}

\subsection{Core Block of COAgents} \label{ss:core_block}
\paragraph{CoreBlock: Gated Graph Convolution Network.}
 We process the immediate neighborhood of each node in the PSG with the Gated Graph Convolutional Network (GGCN)~\cite{GGCN2017,GSP2024}. The GGCN part learns to gate and propagate information along edges of the PSG. It transforms the input representation (details provided in
\textbf{Appendix C1}
 ) into \(\hat{u}^{l}\) and \(\hat{V}^{l}\) (the first line of~\cref{eq:e2e_integration}) using a variant of message passing mechanism.  

\paragraph{CoreBlock: Transformer.}
To capture the long–range dependencies across the PSG, we apply a standard Transformer self‐attention block (the second line in~\cref{eq:e2e_integration}). The M-by-M attention matrix of the transformer encodes (as it is described in \textbf{Appendix C2}) pairwise attention weights across all PSG nodes. Each of the attention weights uniformly scales all the internal features \(\tilde{V_i}\) of a PSG node $i$. 
Since the inner dimension \(N\) (e.g., VRP customers) can be arbitrarily large, we delegate the row‐wise attention over \(\tilde{V_i}\) to problem‐specific modules due to computational overhead. Approximate Transformers implementations, such as Linformer \cite{linformer} or Performer \cite{performer}, can mitigate this computational burden to near-linear complexity. Their application, however, would introduce additional errors in a problem‐dependent manner, making seamless integration into our generic framework challenging without further problem-specific tuning.  

\paragraph{CoreBlock: Problem-Specific End-to-End Modules.}
The problem-specific layers neither possess any intrinsic knowledge about our framework nor operate at the PSG level. Consequently, they can be trivially outsourced from existing literature and updated to enhance framework efficiency as domain-specific SOTA evolves. The E2E module focuses exclusively on the internal structure of the combinatorial problem \(\tilde{V}\) (as it is shown in the last line of~\cref{eq:e2e_integration}), refining it independently from the rest of the framework by leveraging both previous and current local embeddings. It is the framework’s responsibility to adapt to the problem-specific layer and inject meaningful correcting dependencies directly into the latent space of an E2E module. 
\textbf{Appendix C3} lists the architectures that we used for each problem type.

\subsection{Decoders} \label{sec:decoders}
\paragraph{Node \& Move Selection Decoders}
Both the Node Selection Agent and Move Selection Agent use the same decoder head.  Given each node’s final embeddings \((u_i, V_i)\), we first aggregate:
\[
z_i \;=\; \frac{1}{N}\sum_{k=1}^{N}\mathrm{GELU}\bigl(G\,(u_i \concat v_{ik})\bigr),
\]
where \(G\in\mathbb R^{d_{z}\times(d_{u}+d_{v})}\).  Then, letting \(U\in\mathbb R^{1\times d_{z}}\), \(W\in\mathbb R^{1\times(d_{z}+d_{e})}\), and \(\{e_m\}_{m=1}^M\subset\mathbb R^{d_{e}}\) be move embeddings. Denoting \(\sigma\) as the sigmoid function, we compute
\[
p_i \;=\;\sigma\bigl(U\,z_i\bigr),
\qquad
p_{i,m} \;=\;\sigma\bigl(W\,(z_i \concat e_m)\bigr),
\] 

\paragraph{Jump Decoder}
The Jump Decoder decodes the final local embedding \(V_i^{(L)}\in\mathbb{R}^{N\times d_v}\) of a selected solution into a stochastic adjacency matrix \(Y \in \mathbb{R}^{N \times N}\).  Concretely:
\[
Y_{i} = \bigl( \frac{V_{i}^{(L)}W_{Q} W_{K}^\top V_{i}^{(L)\top}}{\sqrt{d_{k}} } \bigr)
\]
We then mask self‐decisions by setting \((Y_i)_{kk}=-\infty\) for all \(k\), and apply a row‐wise softmax:
\[
P_i = \operatorname{softmax}(Y_i)
\;\in[0,1]^{N\times N}.
\]
Each row of \(P_i\) is a probability distribution over decisions (e.g., \ edges or arcs) for one component of the solution. Sampling from \(P_i\) yields a completely new solution in one shot.  In our implementation, we apply this decoder to the local optimal  solution from the previous jump; alternatively, it can be run on any or all PSG nodes to generate diverse candidates.  

\subsection{Training}
\label{sec:training}

Let \(\mathcal D=\{\mathcal G_1,\dots,\mathcal G_N\}\) be our dataset of PSGs.  Each PSG \(\mathcal G\) has a set of solutions \(\mathcal S\) and, for each \(s\in\mathcal S\), a set of candidate moves \(\mathcal M\).

We treat \textit{node and move selection} as two binary classification tasks trained jointly.  For each solution \(s\in\mathcal S\), we assign $y_s \in \{0, 1\}$, which is 1 if $s$ is closest to the optimal solution in and 0 otherwise. For each move \(m\in\mathcal M\), we assign $y_{s,m} \in \{0, 1\}$, which will be 1 if applying $m$ to $s$ yields the best next candidate in $\mathcal{S}$, and 0 otherwise. Given the predicted probabilities \(p_s\) and \(p_{s,m}\) (see section \textit{Decoder}), we minimize the joint binary cross‐entropy:
\begin{equation}
\begin{split}
\mathcal L_{\mathrm{select}}=
-\sum_{i}\Bigl[y_i\log p_i + (1-y_i)\log(1-p_i)\Bigr] \\
-\sum_{i,m}\Bigl[y_{i,m}\log p_{i,m} + (1-y_{i,m})\log(1-p_{i,m})\Bigr].
\end{split}
\end{equation}
At inference, we sample the node index \(i\sim\{p_i\}_i\) and then sample the move \(m\sim\{p_{i,m}\}_m\) according to these distributions.

The \textit{Jump Agent} is trained separately. For each CO instance, we precompute a set of \(K\) target solutions \(\{\delta^{(1)},\dots,\delta^{(K)}\}\), comprising the optimal solution and near‐optimal neighbors that can be converted to optimal with given heuristics.

Let \(\mathcal I\) be the set of all atomic decisions (e.g., \ edges or arcs).  The Jump Decoder outputs probabilities \(P_i\in[0,1]\) for each \(i\in\mathcal I\). For each \(\delta^{(k)}\), we compute labels \(y^{(k)}\in\{0,1\}^{|\mathcal I|}\). Then, we pick the closest reference via $
k^* = \arg\min_k \tfrac12\sum_{i\in\mathcal I}\bigl|y_i^{(k)}-P_i\bigr|,
$ set \(y^*=y^{(k^*)}\), and minimize

\begin{equation}
    \mathcal L_{\mathrm{jump}}
= -\sum_{i\in\mathcal I}\bigl[y^*_i\log P_i + (1-y^*_i)\log(1-P_i)\bigr].
\end{equation}
At inference, decisions \(i\) are sampled according to \(P\) to form a complete solution in one shot.

\paragraph{Training stability.}
Across training, the node and move selection agents exhibited stable convergence because each sampled subgraph contains a single positive move label. The jump agent was more sensitive, since multiple structurally different solutions can be near-optimal and therefore valid jump targets. To reduce label ambiguity, we train the jump model against a set of near-optimal references and select the closest target during training. We did not observe a collapse in the reported runs.

\paragraph{Computational overhead.}
Let $K$ denote the number of retained nodes per PSG subgraph, $|E_{\mathrm{PSG}}|$ the number of PSG edges, $N$ the problem size (e.g., number of customers), $|\mathcal{M}|$ the number of candidate moves, and $d$ the hidden dimension. Each GGCN layer scales as $\mathcal{O}(|E_{\mathrm{PSG}}|\,d)$, the PSG-level Transformer as $\mathcal{O}(K^{2}d)$, the node and move decoders as $\mathcal{O}(Kd + K|\mathcal{M}|d)$, and the jump decoder as $\mathcal{O}(N^{2}d)$ per invocation. The dominant per-iteration costs are therefore PSG-level self-attention, quadratic in $K$, and jump decoding, quadratic in $N$. Because the PSG grows unboundedly with search length, we cap each subgraph at $K \!\le\! 64$ nodes, evicting the oldest entry once the limit is exceeded, while leaving the number of subgraphs in the history pool unrestricted. This bounds the $\mathcal{O}(K^{2}d)$ attention term to a small constant and preserves long-horizon historical context, keeping per-iteration latency low. The $\mathcal{O}(N^{2}d)$ jump-decoder cost remains problem-dependent and is the only term that grows with instance size.

\section{Experimental Setup and Results}
In this section, we evaluate our COAgents framework on two variants of VRP: (1) the Capacitated Vehicle Routing Problem and (2) the Vehicle Routing Problem with Time Windows. Detailed problem definitions are given in \textbf{Appendix D}, and the training data generation procedure is described in \textbf{Appendix E}. All methods are implemented in Python 3.11 using PyTorch 2.4.1 and trained on a machine with six NVIDIA P100-PCIE 16\,GB GPUs under CUDA 12.2. Full reproducibility details, including framework configuration, training data, and per-agent hyperparameters, are provided in \textbf{Appendix F}. Runtimes in Tables~1--2 report total wall-clock time over the full test set (following prior work), whereas Table~3 reports average per-instance inference time.

\subsection{Experimental Test on the VRPTW}
To train our agents, we generate 10K VRPTW instances with 100 customers using a random instance generator. Table \ref{tab:cvrptw_results} compares COAgents against state-of-the-art solvers including heuristics (LKH \cite{lkh3}, HGS \cite{hgs}, OR-Tools) and L2C models (MVMoE \cite{mvmoe}, POMO \cite{pomo}). We evaluate on the 1K test instances introduced by \cite{mvmoe}. All baseline results are taken from \cite{mvmoe} unless otherwise specified. COAgents achieves the best performance across all neural solvers, OR-Tools, and ALNS, reducing the optimality gap by 14\% ($N$=100) and 44\% ($N$=50) relative to the strongest neural solver (POMO), and by 21\% and 40\%, respectively, relative to ALNS.

\begin{table}[htbp!]
\centering
\small
\caption{VRPTW: Average objective (Obj.), gap vs.\ HGS, and runtime on 1k test instances. (H: heuristic, C: learn-to-construct, S: learn-to-search.) Times are total wall-clock time over the full test set.}
\setlength\tabcolsep{3pt}
\renewcommand{\arraystretch}{0.9}
\begin{tabular}{@{}l c| c c c| c c c@{}}
\toprule
\multirow{2}{*}{\textbf{Method}} & \multirow{2}{*}{\textbf{Type}} & \multicolumn{3}{c|}{$N=50$} & \multicolumn{3}{c}{$N=100$}\\
\cmidrule(lr){3-5}\cmidrule(lr){6-8}
 & & Obj. & Gap$\downarrow$ & Time$\downarrow$ & Obj. & Gap$\downarrow$ & Time$\downarrow$\\
\midrule
HGS            & H &14.51 & *        &8.4m   &24.34 & *       &19.6m\\
LKH3           & H &14.61 &0.66\%    &5.5m   &24.72 &1.58\%   &7.8m\\
OR-Tools       & H &14.92 &2.69\%    &10.4m  &25.89 &6.30\%   &20.8m\\
ALNS (1k)      & H &14.97 &2.79\%    &13.4h  &25.49 &4.68\%   &7.3d\\
\midrule
POMO           &C  &14.94 &2.99\%    &3s     &25.37 &4.31\%   &11s\\
POMO-MTL       &C  &15.03 &3.64\%    &3s     &25.61 &5.31\%   &11s\\
MVMoE/4E       &C  &15.00 &3.41\%    &4s     &25.51 &4.90\%   &12s\\
MVMoE/4E-L     &C  &15.01 &3.50\%    &3s     &25.52 &4.93\%   &11s\\
\midrule
\textbf{COAgents} &S &\textbf{14.77} &\textbf{1.67\%}   &3.5h   &\textbf{25.26} &\textbf{3.69\%}   &5.3h\\
\bottomrule
\end{tabular}
\label{tab:cvrptw_results}
\end{table}

\subsection{Experimental Test on CVRP}
We used the 10K CVRP instances with 100 customers introduced by \cite{NeuOpt} to train our agents. Table \ref{tab:cvrp_results} benchmarks our COAgents against the main families of neural VRP solvers on 10k test instances introduced in \cite{NeuOpt}:  
1. \textit{L2P (learning-to-predict):} CVAE-Opt-DE \cite{cvae}, DPDP \cite{dpdp} (state-of-the-art)  
2. \textit{L2C (learning-to-construct):} AM+LCP \cite{am-lcp}, POMO \cite{pomo}, Sym-NCO \cite{symNCO}, POMO+EAS(+SGBS) \cite{pomo-eas,simBeam}  
3. \textit{L2S (learning-to-search):} NLNS~\cite{NeuALNS}, NCE~\cite{nce}, Wu et al.~\cite{wu}, DACT \cite{DACT} (state-of-the-art). Unless stated otherwise, all baseline numbers are taken from \cite{NeuOpt}.  For a fair comparison, we evaluate inference on a single GPU/CPU using a neural processing unit comparable to an A100 GPU \cite{NeuOpt}. Following prior work \cite{pomo-eas,DACT,NeuOpt}, we report:  
(i) average tour cost,  
(ii) optimality gap relative to the state-of-the-art solver HGS \cite{hgs}, and  
(iii) total wall-clock time. 

\begin{table*}[htbp!]
\centering
\small
\caption{CVRP: Average Obj., gap vs.~HGS, and runtime on 10k test instances. (H: heuristic, C: learn-to-construct, S: learn-to-search, P: learn-to-predict, RL: reinforcement learning, SL: supervised learning, UL: unsupervised learning, DE: dynamic encoding, DP: dynamic programming, AS: active search, BS: beam search.). Times are total wall-clock time over the full test set.}
\setlength\tabcolsep{4pt}
\renewcommand{\arraystretch}{1.0}
\resizebox{\textwidth}{!}{%
\begin{tabular}{@{}l c c| c c c| c c c| c c c@{}}
\toprule
\multirow{2}{*}{\textbf{Method}} & \multirow{2}{*}{\textbf{Type}} & \multirow{2}{*}{\textbf{Post}} &
\multicolumn{3}{c|}{$N=20$} & \multicolumn{3}{c|}{$N=50$} & \multicolumn{3}{c}{$N=100$}\\
\cmidrule(lr){4-6}\cmidrule(lr){7-9}\cmidrule(lr){10-12}
 & & & Obj. & Gap$\downarrow$ & Time$\downarrow$ & Obj. & Gap$\downarrow$ & Time$\downarrow$ & Obj. & Gap$\downarrow$ & Time$\downarrow$ \\
\midrule
HGS \cite{hgs} & H & – & 6.13 & * & 2.3m & 10.37 & * & 15m & 15.56 & * & 4.2h \\
LKH3 \cite{lkh3} & H & – & 6.14 & 0.08\% & 4.7m & 10.38 & 0.09\% & 35m & 15.65 & 0.54\% & 9.8h \\
ALNS(1k) \cite{johnn2024graph} & H & – & 6.19 & 1.09\% & 1.3d & 10.78 & 3.93\% & 5.5d & 16.54 & 6.01\% & 18.3d \\
\midrule
CVAE-Opt-DE & P/UL & DE & 6.14 & 0.16\% & 1.5h & 10.40 & 0.33\% & 6.1h & – & – & – \\
DPDP(1M) & P/SL & DP & – & – & – & – & – & – & 15.63 & 0.41\% & 1.2d \\
\midrule
AM+LCP & C/RL & – & 6.15 & 0.33\% & 23m & 10.52 & 1.48\% & 52m & 16.00 & 2.81\% & 2.1h \\
Sym-NCO & C/RL & – & – & – & – & – & – & – & 15.67 & 0.70\% & 7.2h \\
POMO & C/RL & – & 6.14 & 0.09\% & 1.7m & 10.40 & 0.30\% & 11m & 15.67 & 0.70\% & 7.2h \\
POMO+EAS & C/RL & AS & 6.13 & 0.04\% & 6.8m & 10.38 & 0.13\% & 38m & 15.61 & 0.30\% & 16h \\
POMO+EAS+SGBS(short) & C/RL & AS+BS & – & – & – & – & – & – & 15.59 & 0.15\% & 1d \\
POMO+EAS+SGBS(long) & C/RL & AS+BS & – & – & – & – & – & – & 15.58 & 0.10\% & 4.1d \\
\midrule
NLNS(5k) & S/RL & – & 6.18 & 0.73\% & 12m & 10.51 & 1.35\% & 48m & 15.92 & 2.26\% & 2.4h \\
NCE(CROSS) & S/SL & – & 6.13 & 0.00\% & 11h & 10.41 & 0.42\% & 2.3d & 15.81 & 1.59\% & 10.4d \\
Wu et al. & S/RL & – & – & – & – & 10.54 & 1.72\% & 4.2h & 16.17 & 3.87\% & 5h \\
DACT & S/RL & – & 6.13 & 0.01\% & 3m & 10.38 & 0.16\% & 16h & 15.74 & 1.11\% & 1.7d \\
NeuOpt(D2A=1,1k) & S/RL & – & 6.13 & 0.03\% & 2m & 10.43 & 0.61\% & 12m & 15.87 & 1.94\% & 28m \\
\midrule
\textbf{COAgents(1k)} & S/SL & – & 6.18 & 0.34\% & 22.2h & 10.60 & 1.44\% & 1.2d & 16.05 & 2.72\% & 1.9d \\
\bottomrule
\end{tabular}%
}
\label{tab:cvrp_results}
\end{table*}

\subsection{Comparison with ALNS}

In this section, we compare against a strong ALNS baseline that includes all local improvement moves available to COAgents, together with 27 additional destroy-and-repair pairs from \cite{johnn2024graph}. This makes the comparison conservative: ALNS has access to a larger pool of handcrafted operators, while COAgents differs mainly in its learned node selection, move selection, and jump policies. Both algorithms are run for 1000 iterations, and we evaluate the quality of their best-found objective values at ten checkpoints during the search on the 1k VRPTW test set from \cite{mvmoe}. COAgents consistently outperforms ALNS. Despite using the same local improvement heuristics, COAgents outperforms ALNS by replacing static destroy-and-repair operators with a learned Jump Agent. When combined with the NSA’s solution selection and the MSA’s operator choice, this learned diversification provides far more robust and effective guidance than ALNS’s handcrafted policies.

\begin{figure}[htp]
    \centering
    \includegraphics[width=1\columnwidth]{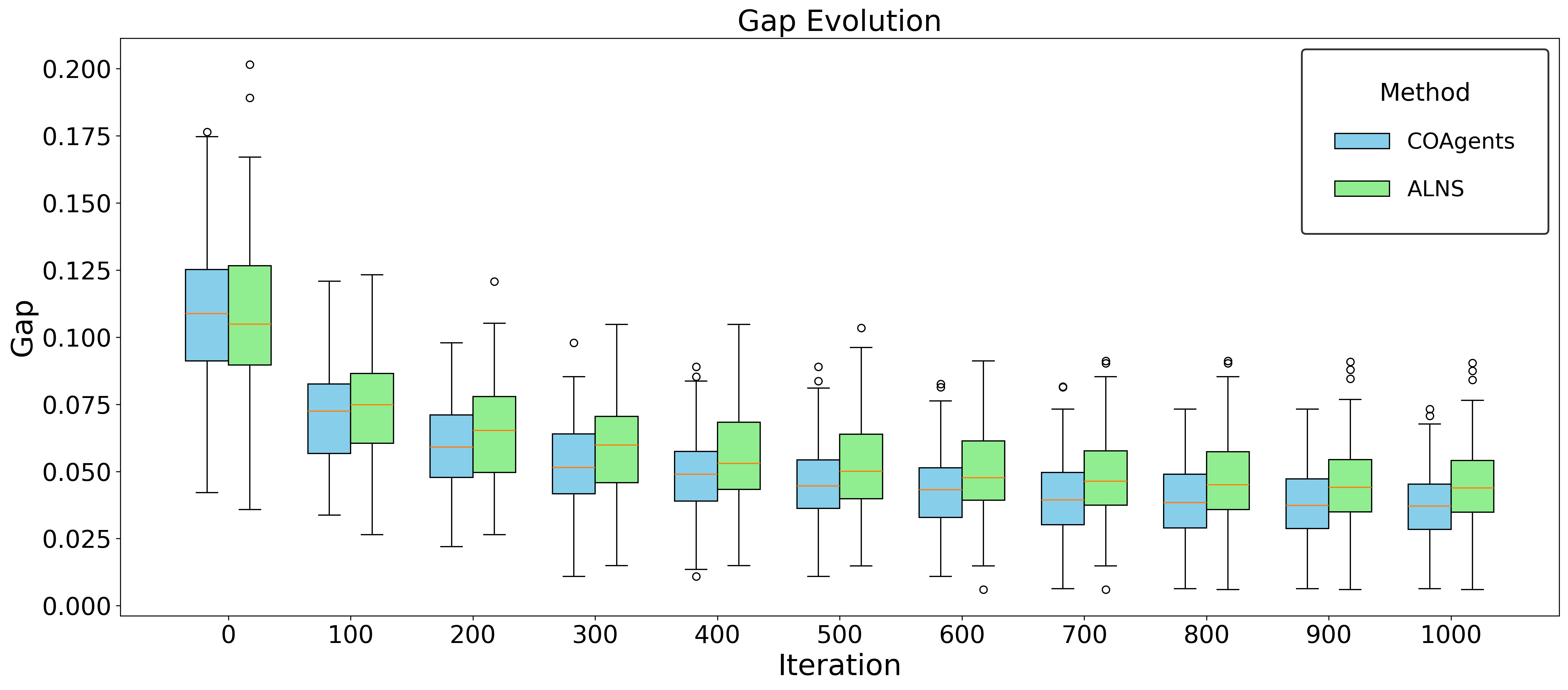}
    \caption{Comparison of optimality gap for a test set of VRPTW instances.}
    \label{fig:gap_evolution_vrptw}
\end{figure}

\subsection{Ablation Study}

To assess the contribution of each component in COAgents, we conducted an ablation study on VRPTW instances. We evaluated three variants: removing the node selection agent (fallback to hill climbing), removing the move selection agent (fallback to the standard ALNS adaptive mechanism), and removing the jump agent. Table~\ref{tab:ablation} reports the average gap to HGS solutions and runtime.
Removing the Jump Agent nearly doubles the gap (from 3.7\% to 7.8\%), making it the most impactful component of COAgents. Nonetheless, the other agents also contribute to the algorithm’s success.

\begin{table}[h]
\centering
\caption{Ablation study on VRPTW instances.}
\label{tab:ablation}
\begin{tabular}{lcc}
\toprule
\textbf{Variant} & \textbf{Avg. Gap} $\downarrow$ & \textbf{Avg. Time, s} $\downarrow$ \\
\midrule
\textbf{COAgents (full)} & \textbf{3.691\%} & \textbf{19} \\
w/o Node-Selection       & 3.705\% & 19 \\ 
w/o Move-Selection       & 4.248\% & 19 \\
w/o Jump Agent           & 7.762\% & 52 \\
\bottomrule
\end{tabular}
\end{table}

\section{Discussion and Future Directions}

Our empirical results demonstrate that COAgents is a compelling alternative to classical heuristics as well as learn-to-construct and learn-to-search methods. On the VRPTW (Table~\ref{tab:cvrptw_results}), COAgents outperforms all neural baselines, OR-Tools, and ALNS, narrowing the gap to HGS by 14--44\% over the strongest neural solver and by 21--40\% over ALNS across N=50 and N = 100.
On the CVRP (Table~\ref{tab:cvrp_results}), it remains competitive with learn-to-search baselines despite relying solely on a compact, problem-specific domain-encoding module. Moreover, in every case COAgents surpasses a pure ALNS implementation, which uses the same local improvement heuristics plus 27 additional destroy–repair pairs, demonstrating that our multi-agent system delivers more robust and effective guidance than ALNS’s handcrafted policies.

\noindent\textbf{Strengths.}
We attribute our superior performance on harder problems, such as the VRPTW, to the PSG, which persistently stores high-quality solutions and feeds them back into our agents. By using this historical information, our agents can drive targeted intensification or diversification more efficiently. In contrast, existing learning-to-search approaches make all their decisions solely based on the current state and their policy weights \cite{johnn2024graph,lagos2024multi}, without ever exploiting accumulated search history.
Furthermore, COAgents' modular design isolates domain-specific logic in the E2E blocks, so transferring to a new problem requires only swapping or tuning that module while the GGCN–Transformer core stays unchanged.

\noindent\textbf{Limitations.} COAgents' runtime is not as competitive as some baseline methods. This is due to two factors: (1) COAgents uses an iterative improvement search that repeatedly generates and modifies solutions, making it slower than constructive methods (i.e,. L2C) that build a solution only once, and (2) our heuristic implementations are in Python, whereas traditional heuristics such as LKH3 and HGS are implemented in C and C++, respectively. However, our execution time is comparable to L2S methods and, in some cases (e.g., NCE), even faster. 
The current implementation is evaluated up to \(N=100\). Scaling to substantially larger or strict real-time VRP settings remains open. The main bottlenecks are PSG-level attention, which grows quadratically with the retained PSG size, and the jump decoder, which constructs an \(N \times N\) decision matrix. Sparse PSG attention, more aggressive history pruning, batched move evaluation, and lighter jump decoders are promising directions for reducing this overhead.

\begin{credits}

\subsubsection{\discintname} The authors have no competing interests to declare that are relevant to the content of this article.

\end{credits}

\bibliographystyle{splncs04}
\bibliography{bibliography}

\clearpage
\appendix
\section{Problem-Specific Feature Definitions}

Below we summarize the global and local features used for each problem domain in our experiments.

\paragraph{Vehicle Routing Problem variants}

Global feature vector \(x_u\) for each CVRP solution includes:
\begin{itemize}
  \item Objective value of the solution.
  \item Number of vehicles.
  \item Number of customers.
  \item Vehicle capacity.
  \item Improvement in objective relative to the parent solution in the PSG.
  \item Sum of objective values of offspring solutions (graph-level aggregate).
  \item Sum of squares of objective values of offspring solutions (graph-level aggregate).
  \item Number of offspring solutions (graph-level aggregate).
\end{itemize}

Local structural matrix \(x_V\) for each VRP solution is a matrix of customer coordinates, demand, time windows (if applicable), and their positions in the route, encoded using cyclic positional encodings \cite{DACT}.

\section{Solutions Positional Embeddings}\label{app:positional}

\paragraph{Positional Feature Embedding for CVRP and VRPTW.} 
Cyclic positional embeddings \cite{DACT} were adopted to implicitly encode the edges of VRP solution. For each dimension $d$ with frequency $\omega_d$,
\begin{equation}
v_{pos} =
\begin{cases}
\sin\!\Bigl(\omega_d \cdot \Bigl|\Bigl(z(i) \bmod \frac{4\pi}{\omega_d}\Bigr)-\frac{2\pi}{\omega_d}\Bigr|\Bigr), & \text{if } d \text{ is even}, \\[1mm]
\cos\!\Bigl(\omega_d \cdot \Bigl|\Bigl(z(i) \bmod \frac{4\pi}{\omega_d}\Bigr)-\frac{2\pi}{\omega_d}\Bigr|\Bigr), & \text{if } d \text{ is odd}.
\end{cases}
\label{embeddings:2}
\end{equation}
where $z(i)$ represents an evenly spaced position for the nodes over a cycle.

\section{Architecture of a Core Block}

\subsection{CoreBlock: GGCN} 

Define the outgoing neighbor set \(\mathcal N^+(i)\) and incoming set \(\mathcal N^-(i)\).

First, we merge global and local features and compute the \textit{aggregated solution embeddings} as follows
\[
z_i^{(l)}
= \frac{1}{N} \sum_{k=1}^{N}
\mathrm{GELU}\bigl(\mathbf{G}\,(u_i^{(l-1)} \concat v_{ik}^{(l-1)})\bigr) \quad \in \mathbb{R}^{d_{z}}
\]
where N is the size of a problem and \(G \in \mathbb{R}^{d_{z} \times (d_{u} + d_{v}})\).
Next, we compute edge‐conditioned gates in both directions:
\begin{align*}
& \mathrm{gate}_{ij}^{(l)} = \\ 
& \quad \tanh\!\Bigl(\frac{1}{|\mathcal N^+(i)|}
\sum_{j\in\mathcal N^+(i)}
\bigl(\mathbf{W}_{K}\,z_i^{(l)} + \mathbf{W}_{Q}\,z_j^{(l)} + \mathbf{W}_{E}\,e_{ij}\bigr)\Bigr),
\\
& \mathrm{gate}_{ji}^{(l)} = \\
& \quad \tanh\!\Bigl(\frac{1}{|\mathcal N^-(i)|}
\sum_{j\in\mathcal N^-(i)}
\bigl(\mathbf{W}_{K}\,z_j^{(l)} + \mathbf{W}_{Q}\,z_i^{(l)} + \mathbf{W}_{E}\,e_{ji}\bigr)\Bigr),
\end{align*}
where \(\mathbf{W}_{K},\mathbf{W}_{Q},\mathbf{W}_{E}\) are learnable keys, queries and \(e_{ij}\) encodes the heuristic on edge \((i\!\to\!j)\).

We then update each hidden feature row via gated message passing:
\begin{align*}
h_{ik}^{(l)}
= & h_{ik}^{(l-1)}
+ \frac{1}{|\mathcal N^+(i)|}\sum_{j\in\mathcal N^+(i)}
\mathrm{gate}_{ij}^{(l)}\,h_{jk}^{(l-1)} \\
+& \frac{1}{|\mathcal N^-(i)|}\sum_{j\in\mathcal N^-(i)}
\mathrm{gate}_{ji}^{(l)}\,h_{jk}^{(l-1)}.
\end{align*}
Here \(h^{(l-1)} = V^{(l-1)}H\) projects the local matrix via \(H\).

Finally, we fuse these updates back into the node embeddings with residual connections:
$$
\hat{u}_i^{(l)}
= u_i^{(l-1)}
+ \mathbf{W}_{u}\Bigl(z_i^{(l)} + \frac{1}{N}\sum_{k=1}^{N}h_{ik}^{(l)}\Bigr),
$$

$$
\hat{v}_{ik}^{(l)}
= v_{ik}^{(l-1)} + \mathbf{W}_{v}\,h_{ik}^{(l)},
$$
where \(\mathbf{W}_{u}\) and \(\mathbf{W}_{v}\) are learnable projections and \(\hat{u}\) and \(\hat{V}=[\hat{v}_{ik}]\) are output representations of global and local embeddings, respectively.

\subsection{CoreBlock: Transformer} \label{app:core_block_Transformer }
The attention and features matrices are treated separately. The attention is computed from global \(\hat{u}\) and internal \(\hat{V}\) features using condensed solution-wide representations:
\[
z_i^{(l)}
= \frac{1}{N} \sum_{k=1}^{N}
\mathrm{GELU}\bigl(\mathbf{G}\,(\hat{u}_i^{(l-1)} \concat \hat{v}_{ik}^{(l-1)})\bigr) \quad \in \mathbb{R}^{d_z}
\]
\[
A^{(l)}
= \operatorname{softmax}\!\Bigl(\tfrac{(Z^{l-1}Q)\,(Z^{(l-1)}K)^\top}{\sqrt{D}}\Bigr)
\quad\in\mathbb{R}^{N\times N},
\]
where \(Z \in \mathbb{R}^{N \times d_{z}}\) is the matrix of all \(z_{i}\) in a PSG, \( Q \in \mathbb{R}^{D \times d_{z}}\) is the queries matrix and \(K \in \mathbb{R}^{D \times d_{z}}\) is the matrix of keys.

In contrast, internal features \(\hat{V}\) are updated in a more granular manner using the unrolled attention-weighted summation over all \(M\) solutions in the PSG:
\[
\tilde{V}_{\mathrm{att}}^{(l)}
= A^{(l)}\, \otimes \bigl(\hat{V}^{(l-1)}W_{V}\bigr)
\quad\in\mathbb{R}^{M \times N \times d_{v}},
\]
for which \(A \in \mathbb{R}^{M \times 1 \times M}\) is broadcasted over N internal dimensions and \(W_{V} \in \mathbb{R}^{1 \times d_{v} \times d_{v}}\) is just a learnable linear projector into the layers latent space. 

Eventually, the \(\tilde{V}\) is updated in a rather typical manner for transformer layers: 
$$
\tilde{V}^{(l)}
= \hat{V}^{(l-1)} + \tilde{V}_{\mathrm{att}}^{(l)},
$$

$$
\tilde{V}^{(l)}
= \mathrm{LayerNorm}\!\bigl(\tilde{V}^{(l)} + \mathrm{FFN}(\tilde{V}^{(l)})\bigr),
$$
where \(\mathrm{FFN}\) is a two‐layer feed‐forward network with GELU activation, and \(\mathrm{LayerNorm}\) denotes layer normalization. This block aggregates information globally — irrespective of graph distance — while residual connections and normalization stabilize training.

\subsection{CoreBlock: Problem-Specific End-to-End Modules} 
For CVRP, we adopt the dual-aspect Transformer architecture from \cite{DACT}, which combines global instance-level representations with a local GCN/GAN module. 
We use the same for VRPTW.

\subsection{Beam search}
The beam search algorithm is described as \cref{alg:beamsearch}. Constrained beam search operates on the initial set of sequences denoted as $S_{0}$. For each item in a sequence, probabilities $P$ are computed using the \textit{Jump} Agent. Within each route, displacement probabilities are averaged to estimate the route's overall optimality. The three least probable routes are identified and discarded. The remaining routes, retained from $S_{0}$, serve as the starting point for the constrained beam search.

Beam search begins with an empty set of completed candidates $F$. At each iteration, every unassigned unique consumer is systematically added to every candidate sequence in the pool $S$, each of length $|S_{0}| + iteration$. To account for route splits or merges, the depot is also included among the possible consumers. Such extension generates up to $N_{Beam} \times |s \notin \text{Routes}|$ new candidate sequences of length $|S_{0}| + iteration + 1$. To maintain the constant beam search width, $N_{Beam} - |F|$ candidates are randomly selected from this expanded pool, using the probability distribution in $P$ as the sampling weight. When any of the new candidate solution $S_{i}$ incorporates all the consumers, it is transferred from the candidate pool $S$ to the pool of finished solutions $F$.

The iteration process terminates when the objective value of the worst completed solution, $\max{F}$, exceeds the best theoretically achievable score among all incomplete candidates, $\min{S}$. The algorithm returns the solution from $F$ with the most favorable objective value.

\begin{algorithm}
\caption{Constrained Beam Search}
\label{alg:beamsearch}
\begin{algorithmic}
\STATE $S_{0}$ \COMMENT {Initial sequence}
\STATE $Routes \gets \max{P(S_{0})}$ \COMMENT{Most probable routes}
\FOR {$i = 1$ \TO $N_{Beam}$ }
    \STATE $S_{i} \gets \forall s \in S_{0} \quad s \notin Routes$ 
\ENDFOR
\STATE $F \gets \emptyset$
\WHILE{$S \neq \emptyset \text{ and } \min{S} < \max{F}$}
    \STATE $B \gets \emptyset$
    \FOR {$i = 1$ \TO $|S|$ }
        \FOR{$j = 1$ \TO $|Routes|$}
            \STATE $B_{ij} \gets S_{i} \cup Routes_{j}$ 
        \ENDFOR
    \ENDFOR
    \STATE $S \gets \text{Sample}(B, P, N_{\text{Beam}} - |F|)$
    \FOR {$i = 1$ \TO $|S|$}
        \IF {$\forall r \in Routes \quad r \in S_{i}$}
            \STATE $F \gets F \cup S_{i}$
        \ENDIF
    \ENDFOR
\ENDWHILE
\STATE \RETURN $\min{F}$
\end{algorithmic}
\end{algorithm}

\section{Benchmark Problems}
\paragraph{Capacitated Vehicle Routing Problem (CVRP)} is defined as follows. Let \(G=(V,E)\) be a complete graph with vertex set \(V = \{0\}\cup C\), where \(0\) denotes the depot and \(C\) the set of customers. Each customer \(i\in C\) has demand \(\delta_i\), and each arc \((i\to j)\in E\) carries weight \(d_{ij}\), the Euclidean distance between \(i\) and \(j\). A solution is a collection of routes—each route is a cycle beginning and ending at the depot, visiting its assigned customers exactly once, subject to the vehicle capacity constraint. The objective is to minimize the sum of all route distances.

\paragraph{Vehicle Routing Problem with Time Window (VRPTW)} is a common extension of the CVRP. Each customer \(i\) has a time window \([e_i,\,\ell_i]\) that specifies the earliest and latest allowable service times. A vehicle may begin service at \(i\) no earlier than \(e_i\) and no later than \(\ell_i\); otherwise, the delivery is infeasible.

\section{Training Dataset Generation} 
\paragraph{Node \& Move Selection Dataset}
For each problem instance in training set, we first compute a near‐optimal solution $s^*$ using a state‐of‐the‐art solver (e.g., HGS for VRP variants \cite{VIDALHGS,PyVRP}). We then perturb $s^*$ to generate a set $\mathcal{S}^P$ of solutions whose objective values lie within controlled gaps of 0.1 \%, 1 \%, 2 \%, 3 \%, 4 \%, 5\% and 10 \% from $s^*$. To ensure diversity in the perturbed nodes while maintaining the target gap, we add a sufficiently large constant penalty on the decisions present in $s^*$ (denoted $I(s^*)$) to the objective function, and repeatedly apply random moves from the move set $\mathcal{M}$ until the perturbed solution meets the desired gap threshold. Next, we restore the original objective function and apply random moves to each perturbed solution $s^P$, keeping only those moves that successfully improve the objective. We repeat this “perturb-and-improve” process many times per instance to collect a large set of training PSGs.

\begin{figure*}[htp]
    \centering
    \includegraphics[width=0.9\textwidth]{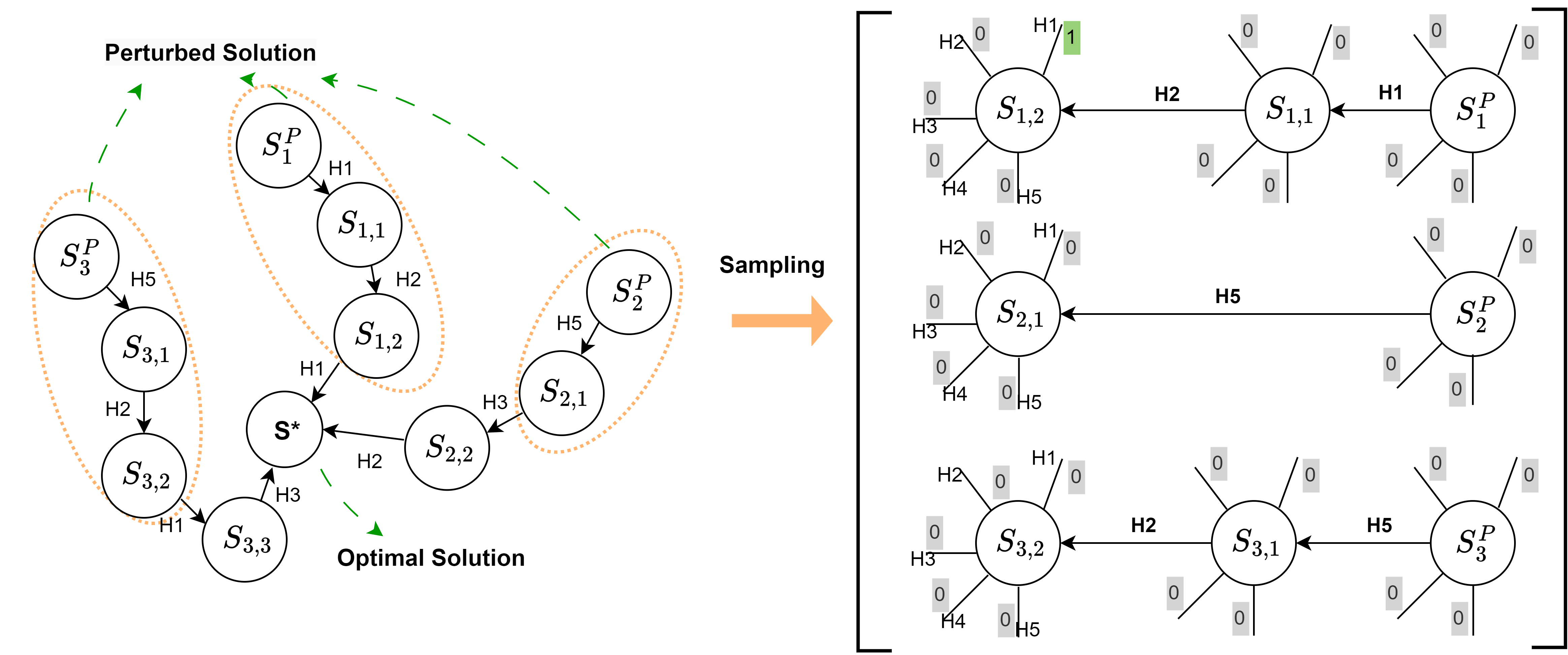}
    \caption{Node \& Move Selection training sample. \textbf{Left:} Three PSGs converging from distinct perturbed solutions. \textbf{Right:} A training sample with three sub-graphs sampled from the PSGs on the left with labels.}
    \label{fig:DatasetGeneration}
\end{figure*}

To form training samples, we then randomly sample subgraphs from these PSGs, selecting at most one move pair from each independent PSG. Let $\mathcal{G}^s = \{\mathcal{G}^s_{1}, \mathcal{G}^s_{2}, \dots, \mathcal{G}^s_{k}\}$ be the set of sampled subgraphs for one problem instance. Within each $\mathcal{G}_i$, we label every candidate move as 0, except for the single move that brings us closest to the original solution $s^*$ that move is labeled 1. Any sample in which multiple moves tie for the best improvement is discarded, ensuring exactly one positive move per subgraph. Fig.~\ref{fig:DatasetGeneration} illustrate an example on how a training sample is generated using the training PSGs.

\paragraph{Jump Agent Dataset} For our jump model, using the same training instances, we generate multiple trajectories per instance by running ALNS \cite{johnn2024graph} from different random starts until convergence to a local optimum. We record each trajectory as a PSG (a sequence of solutions $s_k$ and moves $m$). Then, we sample a subgraph $\mathcal{G}$ as in the node–move dataset and pair each solution $s_k \in \mathcal{G}^s$ with the known global optimum as the prediction target.

\section{Reproducibility}
\label{app:hyper_parameters}

This appendix documents the configuration used to produce all results in the main paper. Framework-level settings shared across agents are listed in Table~\ref{tab:framework_settings}; per-agent training hyperparameters appear in Table~\ref{tab:hyperparams}. Both the Selection Agents and the Jump Agent use a step learning-rate scheduler with step size $100$ and $\gamma = 0.998$. All values are held fixed across VRPTW and CVRP.

\begin{table}[h]
\centering
\caption{Framework-level configuration shared across all agents.}
\label{tab:framework_settings}
\small
\begin{tabular}{ll}
\toprule
\textbf{Component} & \textbf{Setting} \\
\midrule
PSG subgraph cap          & $64$ nodes \\
Pruning rule              & FIFO; discard oldest node on overflow \\
History pool size         & Unbounded \\
Inference budget          & $1{,}000$ iterations \\
\midrule
VRPTW training set        & $10$K synthetic instances, $100$ customers \\
CVRP training set         & $10$K instances from the NeuOpt benchmark \\
\midrule
Hardware                  & $6\times$ NVIDIA P100-PCIE (16\,GB) \\
\bottomrule
\end{tabular}
\end{table}

\begin{table}[h]
\centering
\caption{Training hyperparameters for the Node/Move Selection Agents and the Jump Agent.}
\label{tab:hyperparams}
\small
\begin{tabular}{lcc}
\toprule
\textbf{Hyperparameter} & \textbf{Selection Agents} & \textbf{Jump Agent} \\
\midrule
\multicolumn{3}{l}{\emph{Optimization}} \\
Initial learning rate              & $10^{-4}$ & $10^{-4}$ \\
LR scheduler (step size, $\gamma$) & $(100,\,0.998)$ & $(100,\,0.998)$ \\
Batch size                         & $48$       & $16$ \\
Training iterations                & $50$K      & $25$K \\
\midrule
\multicolumn{3}{l}{\emph{CoreBlock --- GGCN}} \\
Hidden dimension                   & $128$ & $128$ \\
\midrule
\multicolumn{3}{l}{\emph{CoreBlock --- Transformer}} \\
Attention heads                    & $16$  & $16$ \\
Hidden dimension                   & $256$ & $256$ \\
\midrule
\multicolumn{3}{l}{\emph{Decoder}} \\
Attention heads                    & ---   & $4$ \\
\bottomrule
\end{tabular}
\end{table}

\end{document}